\documentclass{article}

\usepackage[english]{babel}

\usepackage[letterpaper,top=2cm,bottom=2cm,left=3cm,right=3cm,marginparwidth=1.75cm]{geometry}

\usepackage{amsmath}
\usepackage{graphicx}
\usepackage[colorlinks=true, allcolors=blue]{hyperref}
\usepackage{amsfonts}
\usepackage{amssymb}
\usepackage{float} 
\usepackage{url}
\usepackage{breqn}
\usepackage{amsthm}

\title{UFT: Unifying Fine-Tuning of SFT and RLHF/DPO/UNA through a Generalized Implicit Reward Function}

\author{
    $\textbf{Zhichao Wang}^{\dagger*1}, \textbf{Bin Bi}^{\dagger1}$, \textbf{Zixu Zhu}$^1$, \textbf{Xiangbo Mao}$^1$, \textbf{Jun Wang}$^1$ \\
    \textbf{Shiyu Wang}$^1$, \textbf{Cheng Wan}$^2$, \textbf{Dong Nie}$^3$, \textbf{Lingzi Hong}$^4$ \\
    $^1$Salesforce \quad $^2$RadixArk \quad $^3$ChatAlpha AI \quad $^4$University of North Texas \\
    \texttt{zcwang0201@gmail.com}}

\begin{document}
\maketitle
\def\thefootnote{}\footnotetext{$*$: corresponding author; $\dagger$: equal contribution\\
\hspace*{5.5mm}github code: \url{https://github.com/zcw0201/UFT-UNA/new/main}}

\begin{abstract}
By pretraining on trillions of tokens, an LLM gains the capability of text generation. However, to enhance its utility and reduce potential harm, SFT and alignment are applied sequentially to the pretrained model. Because SFT and alignment have different objectives and underlying processes, performance on certain tasks can decline.
To address this, we seamlessly introduce Unified Fine-Tuning (UFT), which integrates SFT and alignment into a single training stage using the same objective and loss functions through an implicit reward function. Our experimental results demonstrate that UFT outperforms SFT on instruction-tuning data alone. Moreover, when combining instruction-tuning data with alignment data, UFT effectively prevents the degradation on some tasks across these two stages and shows a clear advantage over sequentially applying SFT and alignment. This is evident in the significant improvements observed in the \textbf{ifeval} task for instruction-following and the \textbf{truthful} task for factuality. The proposed general fine-tuning framework UFT establishes an effective and efficient paradigm for LLM post-training.
\end{abstract}

\section{Introduction}
To enable large language models (LLMs) to understand and generate natural language, they are constructed with billions of parameters and pretrained on datasets containing trillions of tokens \cite{openai2024gpt4}. However, several challenges arise after the pretraining stage of LLMs \cite{wang2024comprehensivesurveyllmalignment}. One major issue is that pretrained LLMs can only continue generation based on the previous context and often struggle to accurately answer users' questions.
To address it, supervised fine-tuning (SFT) is introduced, using pairs of questions and answers. For example, in models like Mistral, preset instructions such as `[INST]' and `[/INST]' are used to frame a question as a prompt \cite{jiang2023mistral7b}. The corresponding answer is then used as the target output. The model's probability of generating the correct answer is maximized through next-token prediction, employing the cross-entropy loss function to classify tokens across the entire token space.

\begin{figure}
\centering
\includegraphics[width=0.95\textwidth]{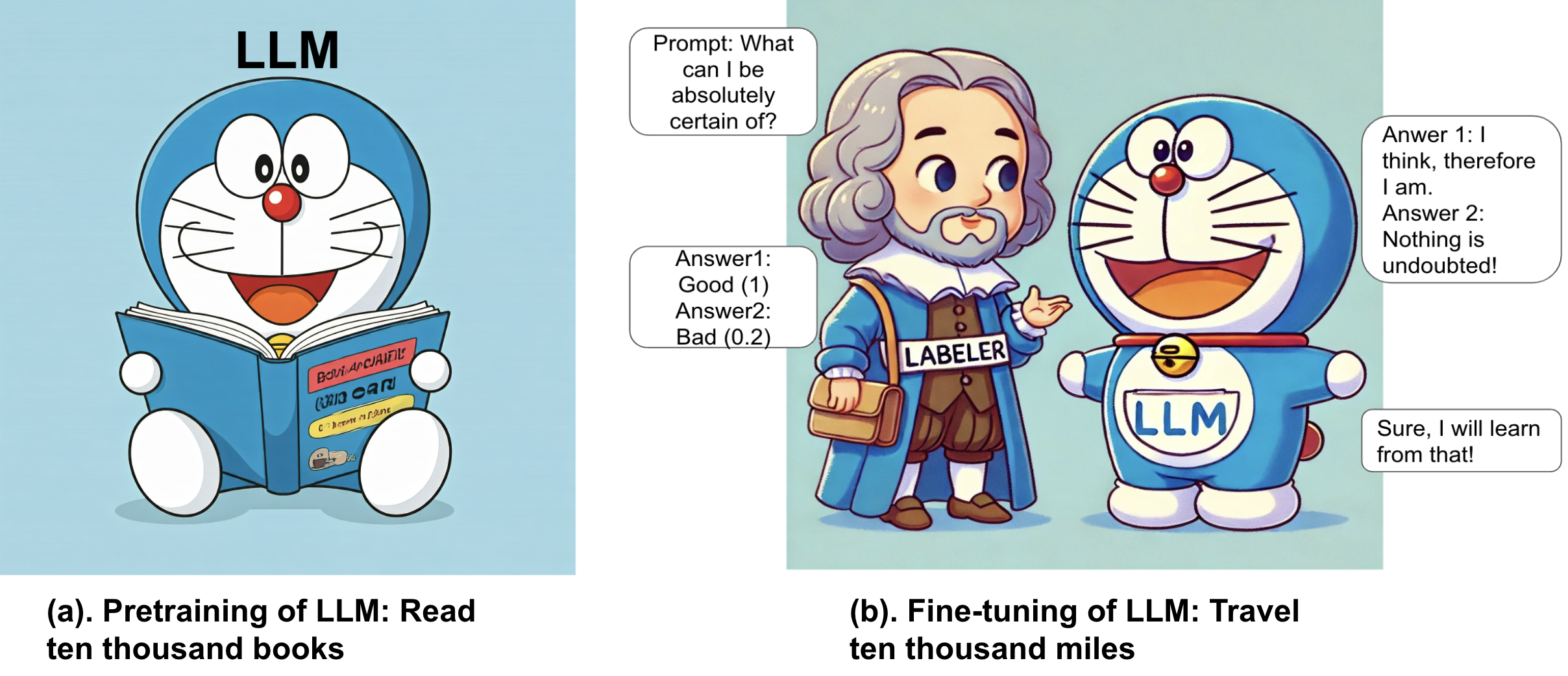}
\caption{UFT integrates SFT and alignment through a generalized implicit reward function. It likens pre-training and fine-tuning of LLMs to Chinese proveb "Read ten thousand books, travel ten thousand miles". In pre-training, the LLM processes vast amounts of data without feedback, gaining broad language understanding. In fine-tuning, it generates responses to prompts and receives feedback, refining its abilities and improving performance on specific tasks.}
\label{fig: proveb}
\end{figure}

The next challenge for LLMs lies in ethical concerns, where LLMs may inadvertently teach humans to engage in unethical activities, such as robbing banks \cite{ouyang2022training}. To address this issue, various alignment methodologies have been proposed, including Reinforcement Learning from Human Feedback (RLHF) \cite{ouyang2022training, bai2022training} with Proximal Policy Optimization (PPO) \cite{schulman2017proximal}, Direct Preference Optimization (DPO) \cite{rafailov2023direct}, Kahneman \& Tversky Optimization (KTO) \cite{ethayarajh2024kto}, and UNified Alignment (UNA) \cite{wang2024unaunifyingalignmentsrlhfppo}. The core idea of alignment is to equip LLMs with the ability to reject harmful requests by learning from human feedback.

For pretrained LLMs, SFT and alignment are traditionally performed in sequence. However, this staged approach often leads to performance degradation, where the model loses capabilities acquired in earlier phases. This paper seeks to address and mitigate this degradation.

Drawing inspiration from UNA’s effectiveness in handling score-based feedback, we propose extending UNA to also fulfill the goals of SFT. By converting SFT data into a format compatible with alignment training, we enable the use of a unified objective and loss function. This approach allows for effective fine-tuning of pretrained LLMs while preserving their previously acquired capabilities across tasks.

The contributions of this paper are listed as follows:
\begin{enumerate}
\item Prove that both UNA and SFT maximize the likelihood of the response in instruction-tuning data, and UNA outperforms SFT on downstream tasks when fine-tuning on instruction-tuning data. 
\item UFT that unifies SFT and alignment solves the performance degradation on instruction-following and factuality tasks elegantly and surpasses the performance of the original sequential application order in downstream tasks.
\item UFT builds a unified post-training framework that is parallel to pretraining where the goal lies in generating responses for given prompts, receiving score-based feedback from different labelers and improve its capability on downstream tasks. 
\end{enumerate}

\section{Methodology}
In this section, we will explore the methodologies of SFT, RLHF, DPO, UNA, and the integrated framework UFT.

\subsection{SFT}
The pretrained LLM is limited to either continuing the text or repeating the question, which restricts its usefulness. To address these limitations and enhance the LLM's question-answering capabilities, SFT is applied. The instruction-tuning dataset consists of numerous pairs of prompts (denoted as $x$) and responses (denoted as $y$). Given a prompt $x$, the probability of the pretrained LLM generating the response $y$ is represented as $\pi_\theta(y|x)$. The objective of SFT is to maximize the probability of all response tokens by using cross-entropy loss as shown in Eq. \ref{eq: SFT loss} and Figure. \ref{fig: UFT}(A).

\begin{equation}
\label{eq: SFT loss}
L_{\text{SFT}}(\pi_\theta) = - \displaystyle\log\left(\pi_\theta(y | x)\right)
\end{equation}

Suppose $y$ is composed of $N$ tokens, i.e., $y_1, y_2, \dots, y_N$. Based on Bayes' theorem, $\pi_\theta(y|x)=\prod_{i=1}^{n} \pi_\theta(y_i | x, y_1, \dots, y_{i-1})$. Applying log to Eq. \ref{eq: SFT loss}, the SFT loss can be derived based on each token using cross entropy loss function on all candidate tokens for classification, i.e., $L_{\text{SFT}}(\pi_\theta) = - \sum_{i=1}^{n}\log\left(\pi_\theta(y_i | x, y_1, \dots, y_{i-1})\right)$.

\subsection{RLHF, DPO and UNA}
\begin{figure*}
\centering
\includegraphics[width=0.95\textwidth]{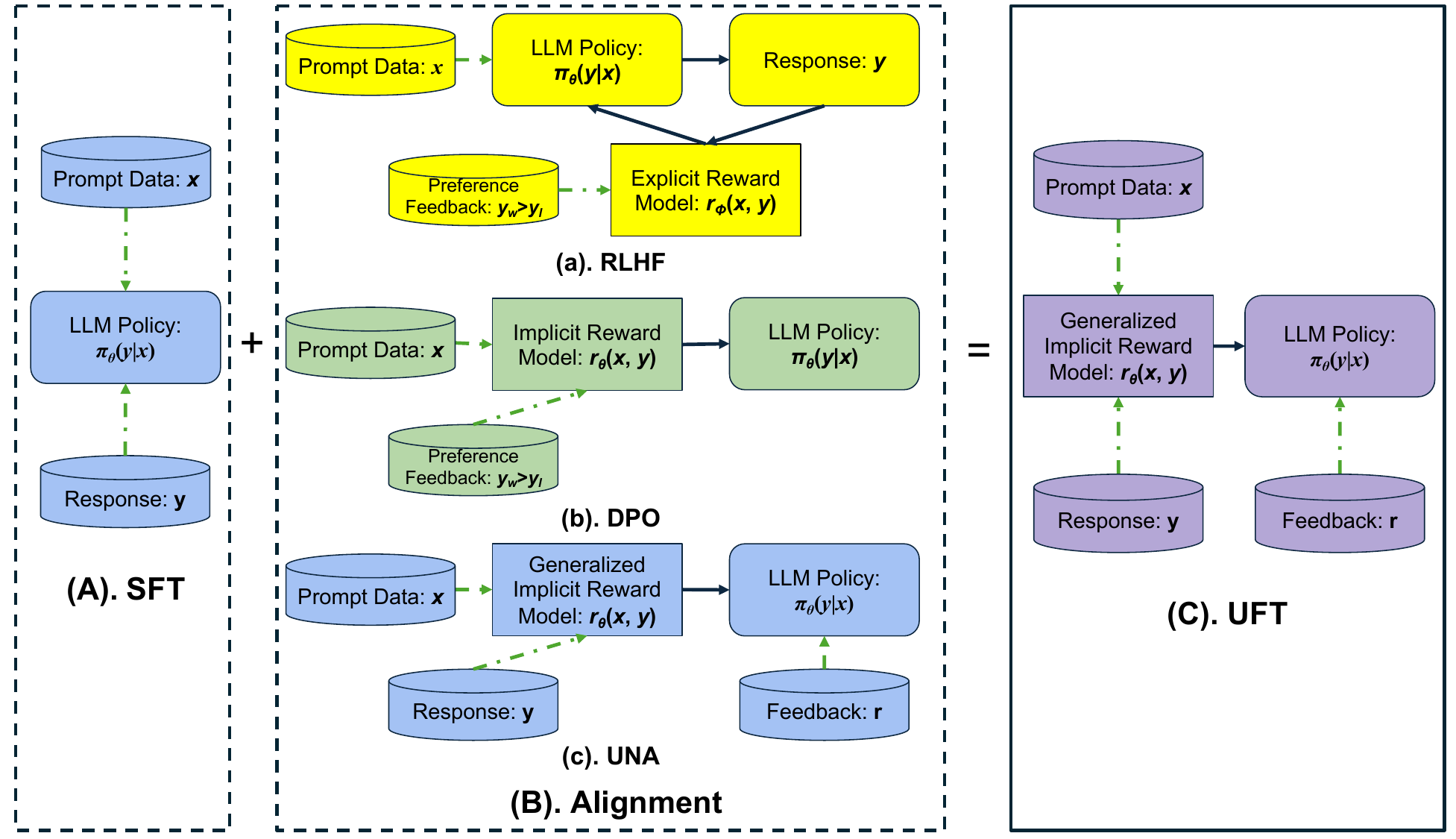}
\caption{(A) refers to SFT, (B) refers to alignment including RLHF, DPO and UNA and (C) refers to UFT. Traditionally, the fine-tuning process begins with SFT followed by alignment. However, the proposed UFT method integrates both SFT and alignment into a single, cohesive process.}
\label{fig: UFT}
\end{figure*}

Even after the pretraining and SFT stages, LLMs can still produce undesired responses that may lead to bias or ethical issues. To address this, methods such as RLHF, DPO, and KTO have been proposed. A plot on the difference among RLHF, DPO and UNA can be found in Figure. \ref{fig: UFT}(B). RLHF tackles these problems in two stages: reward model training and reinforcement learning. During the reward model training process, an explicit reward model is derived from pairwise data using the BT model, as illustrated in Eq. \ref{eq: RM_objective}, where $r_{\phi}$ represents the explicit reward model. The dataset for training the reward model is composed of triplet of 1). prompt $x$, 2). desired response $y_w$ and 3). undesired response $y_l$. The second stage of RLHF involves online reinforcement learning with the pretrained explicit reward model to generate reward signals. These signals are then combined with KL divergence to balance the reward and model capability obtained during the pretraining stage, as shown in Eq. \ref{eq: RL objective}. The RL process is optimized using PPO.

\begin{equation}
\label{eq: RM_objective}
\begin{gathered}
L_{\text{RM}}(\pi_\theta) = - \mathbb{E}_{(x, y_w, y_l) \sim D} \Big[\log\big(\sigma(r_{\phi}(x, y_w)
- r_{\phi}(x, y_l))\big)\Big]
\end{gathered}
\end{equation}

\begin{equation}
\label{eq: RL objective}
\begin{gathered}
\pi^*_\theta(y|x) = \max_{\pi_\theta}\text{E}_{x \sim D} \Big[\text{E}_{y \sim \pi_{\theta}(y | x)} \left[r_\phi(x, y)\right] \\
- \beta D_{\text{KL}} \left( \pi_{\theta}(y|x) \,\|\, \pi_{\text{ref}}(y|x) \right)\Big]
\end{gathered}
\end{equation}

The training of RLHF is memory-intensive because it requires maintaining both the explicit reward model, the value model and the policy model. Additionally, reinforcement learning is notorious for its instability. 

To address this issue, DPO proposes creating a mapping between the optimal policy and the reward model, i.e., an implicit reward model, as illustrated in Eq. \ref{eq: DPO equation}, and optimizing them together. However, $Z(x)$ is intractable and can only be canceled out by subtracting the implicit reward of the desired response $y_w$, i.e., $r_\theta(x, y_w)$ from the implicit reward of the undesired response $y_l$, i.e., $r_\theta(x, y_l)$. This limitation confines DPO to pairwise datasets only. Furthermore, DPO cannot utilize the precise evaluation from the explicit reward model in RLHF. KTO extends DPO to binary feedback by estimating $Z(x)$ from multiple responses to the same prompt.

\begin{equation}
\label{eq: DPO equation}
r_\theta(x, y) = \beta \log \left(\frac{\pi_{\theta}(y|x)}{\pi_{\text{ref}}(y|x)}\right) + \beta \log Z(x)
\end{equation}

To address this issue, UNA offers an alternative proof and demonstrates a novel mapping between the optimal policy and the reward model, i.e., the generalized implicit reward model. The form of the mapping between the generalized implicit reward model and policy is shown in Eq. \ref{eq: UNA optimal reward / policy}. Unlike DPO, which is constrained by $Z(x)$ to pairwise feedback, UNA is versatile enough to handle various types of data, including pairwise feedback, binary feedback, and score-based feedback. These data types are optimized by minimizing the difference between the implicit reward in Eq. \ref{eq: UNA optimal reward / policy} and the explicit reward $r_{\phi}(x, y)$, which is provided by human labelers, other LLMs, or the explicit reward model, as shown in Eq. \ref{eq: general loss for UNA-reward}. When LLMs and explicit reward models are used to evaluate responses in real-time, it is referred to as online UNA. Conversely, if feedback is labeled beforehand, it is termed offline UNA. Consequently, UNA can function in both online and offline modes, effectively bridging the gap between online RLHF and offline DPO and KTO.

\begin{equation}
\begin{split}
\label{eq: UNA optimal reward / policy}
r_{\theta}(x, y) &= \beta \log \left(\frac{\pi_{\theta}(y|x)}{\pi_{\text{ref}}(y|x)}\right)
\end{split}
\end{equation}

\begin{equation}
\begin{split}
\label{eq: general loss for UNA-reward}
L_{\text{UNA}}(\pi_{\theta}) &= \mathbb{E}_{x \sim D, y \sim \pi_{\theta}(\cdot | x)}[g(r_{\phi}(x, y), r_{\theta}(x, y))]
\end{split}
\end{equation}

\subsection{UFT: Unify SFT and Alignment}
\label{UFT: unify SFT and alignment}
To enhance the capabilities of LLMs in question answering and addressing ethical issues, SFT and alignment are typically applied in a sequential manner due to the distinct nature of these tasks. However, this sequential approach often leads to performance degradation on some tasks. In this study, we introduce UFT which integrates SFT and alignment into a single stage, thereby mitigating this problem.

To be more specific, the instruction-tuning data consist of a prompt $x$ and a corresponding response $y$, where the response $y$ is considered of high quality, typically labeled by experts in the field. In comparison, the alignment data include a prompt $x$, response $y$, and feedback $r$. This feedback can be categorized as desired or undesired for pairwise feedback, positive or negative for binary feedback, or a scalar in the interval [0, 1] for general score-based feedback. Due to the high quality of instruction-tuning data, they can be regarded as data with a score of 1, i.e., positive feedback. With this consideration, the instruction-tuning dataset can be transformed into alignment data in the format of prompt $x$, response $y$ and feedback $r=1$, which is the highest reward in consideration. Eventually, the transformed instruction-tuning dataset and the alignment dataset can be merged for fine-tuning LLM using the loss function in UNA as shown in Eq. \ref{eq: general loss for UNA-reward}. Our experiments demonstrate that when fine-tuning using only instruction-tuning dataset, UFT can outperform SFT on downstream tasks, which we attribute to the KL divergence term that focuses on minimization with the pretrained model, a factor ignored in the SFT processes. Additionally, the results indicate that mixing the instruction-tuning data with alignment for UFT can prevent the performance degradation on some tasks and outperform previous sequential methods. Lastly, we discover that the distribution of mixed data, i.e., the proportion of instruction-tuning data and alignment data will impact the performances of LLMs. More details can be found in the experiment section.

A heuristic proof why UFT can replace SFT is provided by arguing that they achieve the same goal of maximizing the probability of $\pi_\theta(y|x)$ for prompt $x$ and response $y$ in the instruction-tuning dataset. For SFT, the probability of $\pi_\theta(y|x)$ is directly maximized by cross entropy-loss. In UFT, suppose we apply the Sigmoid function on the implicit reward in Eq. \ref{eq: UNA optimal reward / policy} and using mean square error (MSE) to measure the difference of implicit reward and explicit reward, the objective will become Eq. \ref{eq: UFT-SFT loss}. This loss function will drive $r_\theta(x,y)=\beta \log \left(\frac{\pi_{\theta}(y|x)}{\pi_{\text{ref}}(y|x)}\right)$ to positive infinity. Since both $\beta$ and $\pi_{\text{ref}}(y|x)$ are fixed, this objective will maximize $\pi_{\theta}(y|x)$. As a result, UFT can replace SFT to maximize the probability of generating the response, and it outperforms SFT by minimizing the difference to the pretrained model.

\begin{equation}
\begin{split}
\label{eq: UFT-SFT loss}
L_{\text{UFT-SFT}}(\pi_{\theta}) = \mathbb{E}_{(x,y) \sim D}\Big[[\sigma(r_{\theta}(x, y))-1]^2\Big] \\
= \mathbb{E}_{(x,y) \sim D}\left\{\left[\sigma\left(\beta \log \left(\frac{\pi_{\theta}(y|x)}{\pi_{\text{ref}}(y|x)}\right)\right)-1\right]^2\right\}
\end{split}
\end{equation}

The generalized UFT loss function is presented in Equation \ref{eq: UFT general loss}. For a given tuple comprising a prompt $x$, response $y$, and explicit reward score $r_{\phi}(x, y)$, the implicit reward $r_{\theta}(x, y)$ is first calculated using Equation \ref{eq: UNA optimal reward / policy}. Subsequently, the difference between the implicit and explicit reward scores is minimized through a general function $g(\cdot)$. Specifically, for Supervised Fine-Tuning (SFT) data, the explicit reward score $r_{\phi}(x, y)$ is set to its maximum value. This approach enables a unified framework for the post-training processes of SFT and alignment.

\begin{equation}
\begin{split}
\label{eq: UFT general loss}
L_{\text{UFT}}(\pi_{\theta}) = \mathbb{E}_{(x,y) \sim D}\Big[ g(r_{\theta}(x, y), r_{\phi}(x, y))\Big]
\end{split}
\end{equation}

In summary, UFT reformats the data structure of instruction-tuning data to ensure compatibility with UNA, thereby enabling the UFT framework to unify SFT and alignment processes. An offline version of UFT is illustrated in part (C) of Figure \ref{fig: UFT}. This offline UFT can be transitioned to an online version, provided that the pre-collected feedback is gathered in real-time using a reward model or other LLMs.

\subsection{Comparison of Pretraining and Fine-Tuning}
By integrating SFT and alignment through UFT, we can establish a unified fine-tuning framework that runs parallel to the pretraining phase. Drawing inspiration from the old Chinese proverb "Read ten thousand books, travel ten thousand miles", we can liken the pretraining stage to "Read ten thousand books". In this phase, a LLM absorbs and learns from an extensive collection of tokens, amounting to trillions. The primary objective during this stage is to predict the next token based on the previous context window, optimizing the model's performance through the cross entropy loss function. Conversely, the fine-tuning stage can be likened to the adage "Travel ten thousand miles," where the model is exposed to a wide array of prompts. It generates diverse responses and learns from the feedback it receives, thereby refining and enhancing its capabilities.

\section{Experiments}
\label{Experiments}
In this section, a series of experiments will be conducted to demonstrate the advantages of UFT across various use cases. Initially, a comparison between UFT and SFT on instruction-tuning data will be performed to illustrate that UFT surpasses SFT. Following this, the second experiment will compare the performance of UFT when applied to SFT and alignment data simultaneously versus applying SFT and alignment sequentially. The third experiment will examine the impact of data distribution between SFT and alignment data, which is crucial for the outcomes of fine-tuning.

\section{Experiment Setup}
To begin with, we compare UFT with SFT on the same instruction-tuning dataset. The UltraChat dataset is utilized by unfolding one conversation into multiple training examples to faciliate the training of UFT. From the UltraChat dataset, 20k samples are selected and utilized for training. The Mistral 7B-v0.1 \cite{jiang2023mistral7b} and Qwen 32B \cite{qwen2025qwen25technicalreport} are utilized as the base model, with low rank adaptation (LoRA) of $r=16$ \cite{hu2021loralowrankadaptationlarge}. For SFT, the learning rate of $1e^{-4}$ is the best, and for UFT, the learning rate of $3e^{-5}$ and $\beta$ of 0.01 is the best.

In UFT vs SFT+Alignment, we utilize the HelpSteer2 dataset of 20k examples for alignment \cite{wang2024helpsteer2opensourcedatasettraining}. For UFT, the 20k examples from UltraChat and 20k examples from HelpSteer2 are merged and utilized for training. For comparison, the best performing SFT model of learning rate $1e^{-4}$ in the previous experiment is utilized for further fine-tuning using DPO, KTO and UNA.

\subsection{UFT vs. SFT}
The fine-tuned models are tested on different tasks including 12 tasks on two HuggingFace Open LLM Leaderboards\cite{open-llm-leaderboard-v1, open-llm-leaderboard-v2}. The new Open LLM Leaderboard includes 6 tasks: \textbf{bbh} (Suzgun et al., 2022), \textbf{gpqa} (Rein et al., 2023), \textbf{mmlu-pro} (Wang et al., 2024), \textbf{musr} (Sprague et al., 2024), \textbf{ifeval} (Zhou et al., 2023), and \textbf{math-hard} (Hendrycks et al., 2021). For all these tasks, the average scores are reported. Conversely, the old Open LLM Leaderboard comprises 6 different tasks: \textbf{gsm8k} (Cobbe et al., 2021), \textbf{truthful} (Lin et al., 2022), \textbf{winograde} (Sakaguchi et al., 2019), \textbf{arc} (Allen AI), \textbf{hellaswag} (Zellers et al., 2019), and \textbf{mmlu} (Hendrycks et al., 2021). In this work, the average match rate in \textbf{gsm8k}, mc2 in \textbf{truthful}, accuracy in \textbf{winograde}, acc-norm in \textbf{arc}, acc-norm in \textbf{hellaswag}, and accuracy in \textbf{mmlu} will be reported. Additionally, MT-Bench \cite{zheng2023judgingllmasajudgemtbenchchatbot} and Alpaca-eval \cite{alpaca_eval} will be used to evaluate the model's ability to generate text responses, rather than selecting from predefined candidate answers. Specifically, the Length-Controlled Win Rate (LC WR) will be documented for Alpaca-eval. Additionally, the average length of the outputs within this evaluation will be provided for comprehensive analysis.

\begin{table*}
\centering
\begin{tabular}{cccccccc}
\hline
\textbf{Model} & \textbf{bbh} & \textbf{gpqa} & \textbf{mmlu-pro} & \textbf{musr} & \textbf{ifeval} & \textbf{math-hard} & \textbf{average} \\
\hline
Mistral & 44.11 & 29.53 & 30.11 & 41.79 & 23.22 & 2.92 & 28.61 \\
\hline
Mistral+SFT & 46.04 & 28.72 & 29.35 & \textbf{42.94} & \textbf{29.5} & 2.66 & 29.87 \\
Mistral+UFT & \textbf{46.55} & \textbf{29.24} & \textbf{30.25} & 41.73 & 28.89 & \textbf{3.87} & \textbf{30.09} \\
\hline
Mistral+SFT+DPO & 44.52 & 29.98 & 29.95 & 40.31 & 26.64 & \textbf{3.13} & 29.09 \\
Mistral+SFT+KTO & 42.89 & 31 & \textbf{30.48} & 40.59 & 25.17 & 2.94 & 28.85 \\
Mistral+SFT+UNA & 43.74 & 30.78 & 30.09 & 40.56 & 26.82 & 2.96 & 29.16 \\
Mistral+UFT & \textbf{45.46} & \textbf{31.15} & 30.05 & \textbf{41.06} & \textbf{46.03} & \textbf{3.13} & \textbf{32.81} \\
\hline
\end{tabular}
\caption{Comparison of Mistral+SFT and Mistral+UFT on instruction-tuning data, and comparison of Mistral+SFT+DPO, Mistral+SFT+KTO, Mistral+SFT+UNA, and Mistral+UFT on both instruction-tuning and alignment data on the new HuggingFace open LLM Leaderboard}
\label{tab:UFT/SFT_new_leaderboard}
\end{table*}

\begin{table*}
\centering
\begin{tabular}{cccccccc}
\hline
\textbf{Model} & \textbf{gsm8k} & \textbf{truthful} & \textbf{winograde} & \textbf{arc} & \textbf{hellaswag} & \textbf{mmlu} & \textbf{average} \\
\hline
Mistral & 38.02 & 42.58 & 77.58 & 61.43 & 83.44 & 62.51 & 60.93 \\
\hline
Mistral+SFT & 39.65 & 51.06 & 78.53 & \textbf{63.99} & \textbf{83.78} & 61.99 & 63.17 \\
Mistral+UFT & \textbf{45.57} & \textbf{51.18} & \textbf{78.93} & 63.82 & 83.54 & \textbf{62.44} & \textbf{64.25} \\
\hline
Mistral+SFT+DPO & 42.19 & 47.83 & 78.45 & 62.16 & 84.03 & \textbf{62.38} & 62.84 \\
Mistral+SFT+KTO & \textbf{42.57} & 49.67 & 79.4 & 61.86 & 83.83 & 62.06 & 63.23 \\
Mistral+SFT+UNA & 39.99 & 49.54 & 79.72 & 62.46 & 84.08 & 62.3 & 63.02 \\
Mistral+UFT & 41.59 & \textbf{54.05} & \textbf{79.79} & \textbf{63.82} & \textbf{84.44} & 62.33 & \textbf{64.34} \\
\hline
\end{tabular}
\caption{Comparison of Mistral+SFT and Mistral+UFT on instruction-tuning data, and comparison of Mistral+SFT+DPO, Mistral+SFT+KTO, Mistral+SFT+UNA, and Mistral+UFT on both instruction-tuning and alignment data on the old HuggingFace open LLM Leaderboard}
\label{tab:UFT/SFT_old_leaderboard}
\end{table*}

\begin{table}
\small
\centering
\begin{tabular}{cccc}
\hline
\textbf{Model} & \textbf{MT-Bench} & \multicolumn{2}{c}{\textbf{Alpaca-eval}} \\
\cline{3-4}
 &  & \textbf{LC WR} & \textbf{Length} \\
\hline
Mistral & 3.15 & 0.31 & 6554  \\
\hline
Mistral+SFT & 6.33 & \textbf{8.07} & 908 \\
Mistral+UFT & \textbf{6.55} & 7.27 & 974 \\
\hline
Mistral+SFT+DPO & 4.81 & 1.05 & 5654 \\
Mistral+SFT+KTO & 4.76 & 0.64 & 6215 \\
Mistral+SFT+UNA & 5.24 & 1.34 & 4945 \\
Mistral+UFT & \textbf{6.78} & \textbf{8.28} & 1317 \\
\hline
\end{tabular}
\caption{Comparison of Mistral+SFT and Mistral+UFT on instruction-tuning data, and comparison of Mistral+SFT+DPO, Mistral+SFT+KTO, Mistral+SFT+UNA, and Mistral+UFT on both instruction-tuning and alignment data using MT-Bench and Alpaca-eval}
\label{tab:UFT/SFT_MTBench_Alpacaeval}
\end{table}

As demonstrated in Table \ref{tab:UFT/SFT_new_leaderboard} and Table \ref{tab:UFT/SFT_old_leaderboard}, Mistral+UFT surpasses Mistral+SFT in 8 out of 12 tasks. When considering the average performances across these leaderboards, Mistral+UFT consistently outperforms Mistral+SFT. This indicates that UFT enhances the capabilities of an LLM by maximizing the reward and minimizing the difference from the pretrained model. Furthermore, in the MT-Bench evaluation, Mistral+UFT outperforms Mistral+SFT, whereas in the Alpaca-eval, Mistral+SFT overshadows Mistral+UFT. From a generation capability standpoint, both UFT and SFT exhibit similar performance. Similar conclusion can be found for Qwen 32B. Qwen+UFT outperforms Qwen+SFT in 10 out of 14 tasks as shown in Table \ref{tab: Qwen UFT/SFT_new_leaderboard} and Table \ref{tab: Qwen UFT/SFT_old_leaderboard} and both MT-Bench and Alpaca-eval in Table \ref{tab: Qwen UFT/SFT_MTBench_Alpacaeval} . As a result, when evaluating the performance on downstream tasks, UFT demonstrates superiority over SFT.

\begin{table*}
\centering
\begin{tabular}{cccccccc}
\hline
\textbf{Model} & \textbf{bbh} & \textbf{gpqa} & \textbf{mmlu-pro} & \textbf{musr} & \textbf{ifeval} & \textbf{math-hard} & \textbf{average} \\
\hline
Qwen & 67.89 & 37.32 & 57.93 & 49.12 & 42.98 & 31.63 & 47.81 \\
\hline
Qwen+SFT & \textbf{67.75} & \textbf{40.42} & \textbf{58.13} & \textbf{48.73} & \textbf{47.48} & 30.77 & \textbf{48.88} \\
Qwen+UFT & 67.67 & \textbf{40.42} & 58.1 & 48.47 & 46.36 & \textbf{31.23} & 48.71 \\
\hline
Qwen+SFT+DPO & 67.8 & 39.01 & 58.06 & \textbf{49.4} & 47.48 & 33.67 & 49.24 \\
Mistral+SFT+KTO & 67.61 & \textbf{39.26} & 58.19 & 48.34 & 53.03 & 35.07 & 50.27 \\
Mistral+SFT+UNA & \textbf{68.84} & 38.42 & 58.35 & 49.14 & 57.33 & 34.7 & 51.13 \\
Mistral+UFT & 67.73 & 38.59 & \textbf{58.73} & 47.42 & \textbf{64.05} & \textbf{37.8} & \textbf{52.39} \\
\hline
\end{tabular}
\caption{Comparison of Qwen+SFT and Qwen+UFT on instruction-tuning data, and comparison of Qwen+SFT+DPO, Qwen+SFT+KTO, Qwen+SFT+UNA, and Qwen+UFT on both instruction-tuning and alignment data on the new HuggingFace open LLM Leaderboard}
\label{tab: Qwen UFT/SFT_new_leaderboard}
\end{table*}

\begin{table*}
\centering
\begin{tabular}{cccccccc}
\hline
\textbf{Model} & \textbf{gsm8k} & \textbf{truthful} & \textbf{winograde} & \textbf{arc} & \textbf{hellaswag} & \textbf{mmlu} & \textbf{average} \\
\hline
Qwen & 85.79 & 57.75 & 82 & 70.31 & 85.22 & 83.19 & 77.38 \\
\hline
Qwen+SFT & 87.38 & 58.81 & 82 & \textbf{70.9} & 84.99 & \textbf{83.39} & 77.91 \\
Qwen+UFT & \textbf{88.29} & \textbf{58.96} & \textbf{82.64} & \textbf{70.9} & \textbf{85} & 83.34 & \textbf{78.19} \\
\hline
Qwen+SFT+DPO & 85.67 & 58.54 & 82 & 70.14 & 85.24 & 83.31 & 77.48 \\
Qwen+SFT+KTO & 85.56 & 59.77 & 82.16 & 70.14 & 85.32 & 83.32 & 77.71 \\
Qwen+SFT+UNA & \textbf{92.08} & 62.74 & 82.4 & 71.5 & 85.37 & \textbf{83.36} & 79.58 \\
Qwen+UFT & 90.68 & \textbf{66.7} & \textbf{83.58} & \textbf{71.84} & \textbf{85.58} & 83.35 & \textbf{80.29} \\
\hline
\end{tabular}
\caption{Comparison of Qwen+SFT and Qwen+UFT on instruction-tuning data, and comparison of Qwen+SFT+DPO, Qwen+SFT+KTO, Qwen+SFT+UNA, and Qwen+UFT on both instruction-tuning and alignment data on the old HuggingFace open LLM Leaderboard}
\label{tab: Qwen UFT/SFT_old_leaderboard}
\end{table*}

\begin{table}
\small
\centering
\begin{tabular}{cccc}
\hline
\textbf{Model} & \textbf{MT-Bench} & \multicolumn{2}{c}{\textbf{Alpaca-eval}} \\
\cline{3-4}
 &  & \textbf{LC WR} & \textbf{Length} \\
\hline
Qwen & 7.97 & 8.47 & 2176 \\
\hline
Qwen+SFT & 7.85 & 8.34 & 2800 \\
Qwen+UFT & \textbf{7.95} & \textbf{9.96} & 1292 \\
\hline
Qwen+SFT+DPO & 8.64 & 9.83 & 2664 \\
Qwen+SFT+KTO & 8.58 & 10.17 & 2149 \\
Qwen+SFT+UNA & 8.57 & 13.75 & 1328 \\
Qwen+UFT & \textbf{8.67} & \textbf{13.79} & 1307 \\
\hline
\end{tabular}
\caption{Comparison of Qwen+SFT and Qwen+UFT on instruction-tuning data, and comparison of Qwen+SFT+DPO, Qwen+SFT+KTO, Qwen+SFT+UNA, and Qwen+UFT on both instruction-tuning and alignment data using MT-Bench and Alpaca-eval}
\label{tab: Qwen UFT/SFT_MTBench_Alpacaeval}
\end{table}

\subsection{UFT vs SFT+Alignment}
The same tasks are utilized for evaluation. In the two HuggingFace open LLM leaderboards, Mistral+UFT outperforms all of Mistral+SFT+DPO, Mistral+SFT+KTO, and Mistral+SFT+UNA in 9 out of 12 tasks. In terms of  average scores, Mistral+UFT surpasses all three sequential methods. Several aspects need further discussion.
Firstly, the performance degradation problem is evident as the performances of Mistral+SFT+DPO, Mistral+SFT+KTO, and Mistral+SFT+UNA are worse than Mistral+SFT, a phenomenon also known as alignment tax. Additionally, we observe that Mistral+UFT shows significant improvements on \textbf{ifeval}, which tests the model's capability of instruction-following, and \textbf{truthful}, which assesses the model's alignment capabilities. These results indicate that UFT greatly enhances instruction-following and alignment capabilities, demonstrating its effectiveness.
In terms of generation capability, Mistral+UFT outperforms the other three sequential methods and it does not seem to suffer from performance degradation, as it performs better than both Mistral+SFT and Mistral+UFT on instruction-tuning data alone. Moreover, Mistral+UFT does not bias towards long generation like the other three sequential methods, which is another advantage of UFT. Simiar observations can be found for Qwen 32B. Qwen+UFT outperforms Qwen+SFT+Alignment in 9 out of 14 tasks, especially the average of the new and old HuggingFace open LLM leaderboards as shown in Table. \ref{tab: Qwen UFT/SFT_new_leaderboard} and Table. \ref{tab: Qwen UFT/SFT_old_leaderboard}. For generation capability, Qwen+UFT outperforms Qwen+SFT+Alignment on both MT-Bench and Alpaca-eval, though the benefits diminish compared with smaller Mistal 7B model. This means that the performance degradation caused by SFT and alignment is diminished with larger model size, and this is consistent the conclusions of previous works on RLHF \cite{bai2022training}.

\begin{table*}
\centering
\begin{tabular}{cccccccc}
\hline
\textbf{Model} & \textbf{bbh} & \textbf{gpqa} & \textbf{mmlu-pro} & \textbf{musr} & \textbf{ifeval} & \textbf{math-hard} & \textbf{average} \\
\hline
Mistral & 44.11 & 29.53 & \textbf{30.11} & \textbf{41.79} & 23.22 & 2.92 & 28.61 \\
UNA(16k+20k) & 45.17 & 30.64 & 29.95 & 39.19 & 44.21 & 2.38 & 31.92 \\
UNA(20k+20k) & \textbf{45.46} & 31.15 & 30.05 & 41.06 & 46.03 & 3.13 & \textbf{32.81} \\
UNA(32k+20k) & 44.75 & 31.28 & 29.76 & 39.06 & \textbf{46.76} & \textbf{3.51} & 32.52 \\
UNA(65k+20k) & 44.4 & \textbf{31.38} & 29.66 & 36.8 & 41.91 & \textbf{3.51} & 31.28 \\
UNA(130k+20k) & 44.35 & 29.75 & 30.02 & 38.66 & 44.2 & 2.99 & 31.66 \\
UNA(260k+20k) & 44.31 & 31 & 30.1 & 39.85 & 45.8 & 3.26 & 32.39 \\
\hline
\end{tabular}
\caption{Impact of different distributions of instruction-tuning and alignment data on the new HuggingFace Open LLM Leaderboard}
\label{tab:UFT_data_distribution_new_leaderboard}
\end{table*}

\begin{table*}
\centering
\begin{tabular}{cccccccc}
\hline
\textbf{Model} & \textbf{gsm8k} & \textbf{truthful} & \textbf{winograde} & \textbf{arc} & \textbf{hellaswag} & \textbf{mmlu} & \textbf{average} \\
\hline
Mistral & 38.02 & 42.58 & 77.58 & 61.43 & 83.44 & 62.51 & 60.93 \\
UNA(16k+20k) & 40.83 & \textbf{56.69} & 79.4 & 64.85 & \textbf{84.66} & 62.22 & \textbf{64.78} \\
UNA(20k+20k) & \textbf{41.59} & 54.05 & 79.79 & 63.82 & 84.44 & 62.33 & 64.34\\
UNA(32k+20k) & 40.83 & 54.38 & 79.72 & 64.42 & 84.46 & \textbf{62.88} & 64.45 \\
UNA(65k+20k) & 41.43 & 52.35 & 79.79 & \textbf{65.27} & 84.19 & 61.89 & 64.15 \\
UNA(130k+20k) & 41.13 & 50.17 & 79.48 & 64.59 & 84.07 & 62.26 & 63.62 \\
UNA(260k+20k) & 41.21 & 50.17 & \textbf{80.43} & 64.85 & 83.89 & 62.35 & 63.82 \\
\hline
\end{tabular}
\caption{Impact of different distributions of instruction-tuning and alignment data on the old HuggingFace Open LLM Leaderboard}
\label{tab:UFT_data_distribution_old_leaderboard}
\end{table*}

\begin{table}
\small
\centering
\begin{tabular}{cccc}
\hline
\textbf{Model} & \textbf{MT-Bench} & \multicolumn{2}{c}{\textbf{Alpaca-eval}} \\
\cline{3-4}
 &  & \textbf{LC WR} & \textbf{Length} \\
\hline
Mistral & 3.15 & 0.31 & 6554  \\
UNA(16k+20k) & 6.25 & 7.9 & 1378 \\
UNA(20k+20k) & 6.78 & 8.28 & 1317 \\
UNA(32k+20k) & 6.54 & 8.23 & 1324 \\
UNA(65k+20k) & 6.3 & \textbf{9.92} & 1338 \\
UNA(130k+20k) & \textbf{6.83} & 7.43 & 1361 \\
UNA(260k+20k) & 6.45 & 6.85 & 1378 \\
\hline
\end{tabular}
\caption{Impact of different distributions of instruction-tuning and alignment data on MT-Bench and Alpaca-eval}
\label{tab:UFT_data_distribution_MTBench_Alpacaeval}
\end{table}

\subsection{Data Distribution in Training UFT}
During the pretraining phase, ensuring a balanced data distribution is crucial for optimizing the capabilities of a LLM. In line with this principle, we have incorporated varying percentages of the instruction-tuning data into the alignment dataset to create a final training dataset. Specifically, we have kept the alignment dataset constant, i.e., 20k while integrating 260k, 130k, 65k, 32k, and 16k samples from the UltraChat dataset. These combined datasets are then used for fine-tuning, allowing us to observe the impact of different data distributions on the model's performance.

Several intriguing points will be discussed. As shown in Table \ref{tab:UFT_data_distribution_new_leaderboard} and Table \ref{tab:UFT_data_distribution_old_leaderboard}, among the 12 tasks, one task, namely \textbf{musr}, shows a minor statistical decrease in performance. In contrast, four tasks-\textbf{gpqa}, \textbf{gsm8k}, \textbf{winograde}, and \textbf{arc}—exhibit statistically small improvements. Notably, two tasks, \textbf{ifeval} and \textbf{truthful}, demonstrate statistically significant improvements. Specifically, \textbf{ifeval} improves from 23.22 to around 44, and \textbf{truthful} improves from 42.58 to around 53.
Significant advancements are observed in MT-Bench and Alpaca-eval, indicating a substantial enhancement in the generation capability of the LLM as shown in Table \ref{tab:UFT_data_distribution_MTBench_Alpacaeval}. This aligns with our expectations, as instruction-tuning data primarily aim to enhance the model's instruction-following capabilities, which positively impacts \textbf{ifeval}, MT-Bench, and Alpaca-eval. Meanwhile, alignment data contribute to improvements in tasks like \textbf{truthful}. Other tasks remain largely unaffected due to the absence of relevant data, suggesting that incorporating more pertinent data could enhance their performance.
When increasing the proportion of instruction-tuningg data, we observe a decrease in \textbf{truthful} performance from 56 to 50, indicating that a larger proportion of alignment data benefits the bias and ethical performance of the LLM. On the other hand, simply adding more instruction-tuning data does not improve the performance of \textbf{ifeval}, MT-Bench, and Alpaca-eval. Therefore, further investigation into the optimal ratio between instruction-tuning data and alignment data is warranted.
\section{Related Work}
The field of large language models (LLMs) has undergone significant advancements, marked by models trained on trillions of tokens and comprising billions of parameters, often processed in parallel during their pretraining phase \cite{openai2024gpt4,anthropic2024claude,team2023gemini}. Subsequently, supervised fine-tuning (SFT) becomes essential for enhancing the model's performance across diverse downstream tasks.

However, neither pretraining nor SFT can fully address the issues of bias and ethics in LLMs \cite{openai2024gpt4, wang2024comprehensivesurveyllmalignment}. To tackle these challenges, RLHF with PPO has been proposed and is widely accepted for aligning LLMs, including GPT and Claude \cite{ouyang2022training, bai2022training}. Despite its popularity, RLHF/PPO faces several issues, such as high memory requirements, instability in reinforcement learning, and the need for multiple training stages, including reward model (RM) training and RL fine-tuning \cite{rafailov2023direct}. To reduce the cost of human labeling, AI feedback can be used to replace human feedback, a method known as reinforcement learning from AI feedback (RLAIF) \cite{bai2022constitutional, lee2023rlaif}. RLOO argues that PPO is excessive for LLM alignment since the model has already been pretrained, suggesting that RLOO should suffice \cite{ahmadian2024basicsrevisitingreinforcestyle}. 

To simplify RLHF, DPO has been proposed to map the optimal policy and reward model into a single step, transforming the initial unstable RL into a binary cross-entropy problem \cite{rafailov2023direct}. Several downstream studies have been conducted on DPO to expand and generalize its capabilities, such as different loss functions \cite{pal2024smaug, azar2023general}, iterative approaches \cite{yuan2024selfrewardinglanguagemodels, xu2024thingscringeothersiterative}, token DPO \cite{rafailov2024rqlanguagemodel, zeng2024tokenleveldirectpreferenceoptimization} and stepwise DPO \cite{kim2024sdpo}.

Previous work focused on pairwise datasets, which are more challenging to gather. In contrast, binary feedback like "thumbs up" and "thumbs down" is easier to collect. KTO leverages the concept of human aversion to undesired data and successfully handles binary feedback \cite{ethayarajh2024kto}. DRO focuses on binary data by estimating the policy and value functions and optimizing each sequentially while keeping the other fixed \cite{richemond2024offlineregularisedreinforcementlearning}. However, these methods cannot accommodate different types of data. To address this, UNA was proposed to handle pairwise, binary, and score-based feedback through an implicit reward model, supporting both online and offline alignment \cite{wang2024unaunifyingalignmentsrlhfppo}. 

There have also been efforts to merge SFT with alignment. ORPO proposed a new loss function to increase the ratio of desired responses over undesired responses, achieving both SFT and alignment \cite{hong2024orpomonolithicpreferenceoptimization}. PAFT suggested conducting SFT and alignment in parallel and merging them afterward \cite{pentyala2024paftparalleltrainingparadigm}. However, ORPO's reliance on pairwise datasets and its deteriorating performance compared to other SFT and alignment methods pose challenges. In contrast, PAFT requires training separate adaptors for SFT and alignment and merging them through sparsity, which is inefficient. Inspired by UNA's achievements, we aim to unify SFT with alignment based on UNA's principles to avoid performance degradation.

\section{Conclusion}
During pretraining, an LLM is trained on vast datasets to predict the next token based on a given context window, enabling it to learn rich linguistic patterns and general knowledge. In contrast, fine-tuning involves exposing the model to specific prompts, generating responses, receiving feedback, and refining its performance on targeted downstream tasks. However, traditional sequential approaches to SFT and alignment can disrupt the model’s pretrained knowledge, leading to a decline in generalization or task-specific performance.

UFT addresses these challenges by streamlining the fine-tuning process. When trained exclusively on instruction-tuning data, UFT outperforms conventional SFT by minimizing divergence from the pretrained model, achieved through careful optimization of KL divergence. This ensures that the model retains its foundational capabilities while adapting to new tasks. Furthermore, by strategically blending instruction-tuning and alignment data, UFT surpasses traditional sequential SFT+alignment methods, effectively mitigating performance degradation across a range of tasks. In summary, UFT establishes a powerful and efficient fine-tuning paradigm that balances task-specific adaptation with the retention of pretrained knowledge, paving the way for more capable and versatile LLMs.

\section{Limitation}
\label{Limitation}
This study presents two primary limitations that warrant attention in future research. First, the analysis is restricted to English, necessitating a comprehensive evaluation of the approach's multilingual capabilities. Second, the current research primarily utilized academic datasets (e.g., Ultrachat and Helpsteer). To enhance the real-world applicability of these findings, further validation on diverse industrial datasets is recommended.

\bibliographystyle{alpha}
\bibliography{sample}

\newpage
\appendix

\section{SFT on 20k examples with Different LR}

The impact of different learning rates on the performances of SFT can be found in Table. \ref{tab:SFT_lr_new_leaderboard} and Table. \ref{tab:SFT_lr_old_leaderboard} for the new and old HuggingFace Open LLM Leaderboards.

\section{UFT on 20k examples with Different LR and $\beta$}

The impact of different learning rates and $\beta$ on the performances of UFT can be found in Table. \ref{tab:UFT_lr_new_leaderboard} and Table. \ref{tab:UFT_lr_old_leaderboard} for the new and old HuggingFace Open LLM Leaderboards.

\begin{table*}
\centering
\begin{tabular}{cccccccc}
\hline
\textbf{LR} & \textbf{bbh} & \textbf{gpqa} & \textbf{mmlu-pro} & \textbf{musr} & \textbf{ifeval} & \textbf{math-hard} & \textbf{average} \\
\hline
$3e^{-6}$ & 45.87 & 29.39 & \textbf{30.21} & 42.41 & 25.2 & \textbf{3.61} & 29.45 \\
$1e^{-5}$ & 47.14 & 29.59 & \textbf{30.21} & 41.74 & 25.77 & 3.12 & 29.6 \\
$3e^{-5}$ & \textbf{47.71} & \textbf{29.76} & 30.1 & 40.95 & 26.96 & 3.01 & 29.75 \\
$1e^{-4}$ & 46.04 & 28.72 & 29.35 & 42.94 & \textbf{29.5} & 2.66 & \textbf{29.87} \\
$3e^{-4}$ & 42.79 & 28.17 & 25.22 & \textbf{43.61} & 19.78 & 2.3 & 26.98 \\
\hline
\end{tabular}
\caption{The impact of different learning rates for SFT on the new HuggingFace Open LLM Leaderboard}
\label{tab:SFT_lr_new_leaderboard}
\end{table*}

\begin{table*}
\centering
\begin{tabular}{cccccccc}
\hline
\textbf{LR} & \textbf{gsm8k} & \textbf{truthful} & \textbf{winograde} & \textbf{arc} & \textbf{hellaswag} & \textbf{mmlu} & \textbf{average} \\
\hline
$3e^{-6}$ & 42.95 & 47.9 & \textbf{79.32} & 61.86 & 83.34 & 62.37 & 62.96 \\
$1e^{-5}$ & 43.25 & 49.58 & 79.08 & 62.29 & 83.33 & 62.15 & 63.28 \\
$3e^{-5}$ & \textbf{44.47} & 50.5 & 78.93 & 62.54 & 83.25 & \textbf{62.61} & \textbf{63.72} \\
$1e^{-4}$ & 39.65 & \textbf{51.06} & 78.53 & \textbf{63.99} & \textbf{83.78} & 61.99 & 63.17 \\
$3e^{-4}$ & 13.39 & 46.43 & 75.85 & 58.53 & 81.16 & 55.73 & 55.18 \\
\hline
\end{tabular}
\caption{The impact of different learning rates for SFT on the old HuggingFace Open LLM Leaderboard}
\label{tab:SFT_lr_old_leaderboard}
\end{table*}

\begin{table*}
\centering
\begin{tabular}{cccccccc}
\hline
\textbf{LR/} $\beta$ & \textbf{bbh} & \textbf{gpqa} & \textbf{mmlu-pro} & \textbf{musr} & \textbf{ifeval} & \textbf{math-hard} & \textbf{average} \\
\hline
$1e^{-4}/0.01$ & \textbf{47.01} & 27.86 & 30.3 & 39.07 & \textbf{32.49} & 3.57 & 30.08 \\
$1e^{-4}/0.03$ & 46.09 & 29.5 & 30.09 & 40.8 & 29.27 & 3.84 & 29.93 \\
$1e^{-4}/0.1$ & 45.57 & 30.5 & 30.15 & 42.42 & 27.21 & 2.94 & 29.8 \\
$1e^{-4}/0.3$ & 44.02 & 30.04 & 29.65 & \textbf{43.76} & 26.08 & 2.64 & 29.37 \\
\hline
$3e^{-5}/0.01$ & 46.55 & 29.24 & 30.25 & 41.73 & 28.89 & \textbf{3.87} & \textbf{30.09} \\
$3e^{-5}/0.03$ & 46.56 & 28.72 & 30.44 & 40.67 & 29.7 & 3.08 & 29.86 \\
$3e^{-5}/0.1$ & 46.31 & 29.29 & \textbf{30.7} & 40.95 & 26.52 & 3.5 & 29.55 \\
$3e^{-5}/0.3$ & 44.91 & 28.6 & 30.44 & 43.1 & 26.54 & 3.1 & 29.45 \\
\hline
$1e^{-5}/0.01$ & 46.44 & 29.27 & 30.44 & 38.15 & 26.02 & 3.59 & 28.99 \\
$1e^{-5}/0.03$ & 46.46 & 29.51 & 30.54 & 39.21 & 25.39 & 3.02 & 29.02 \\
$1e^{-5}/0.1$ & 45.8 & 29.99 & 30.41 & 42.96 & 26.67 & 3.01 & 29.81 \\
$1e^{-5}/0.3$ & 45.03 & 29.74 & 30.46 & 42.97 & 25 & 2.57 & 29.3 \\
\hline
$3e^{-6}/0.01$ & 45.83 & 29.86 & 30.5 & 41.5 & 25.42 & 2.8 & 29.32 \\
$3e^{-6}/0.03$ & 45.81 & 29.91 & 30.45 & 41.24 & 25.23 & 2.58 & 29.2 \\
$3e^{-6}/0.1$ & 45.48 & \textbf{30.61} & 30.53 & 43.09 & 23.87 & 3.27 & 29.48 \\
$3e^{-6}/0.3$ & 44.3 & 29.48 & 30.29 & 42.71 & 22.56 & 3.06 & 28.73 \\
\hline
\end{tabular}
\caption{The impact of different learning rates for SFT on the new HuggingFace Open LLM Leaderboard}
\label{tab:UFT_lr_new_leaderboard}
\end{table*}

\begin{table*}
\centering
\begin{tabular}{cccccccc}
\hline
\textbf{LR/} $\beta$ & \textbf{gsm8k} & \textbf{truthful} & \textbf{winograde} & \textbf{arc} & \textbf{hellaswag} & \textbf{mmlu} & \textbf{average} \\
\hline
$1e^{-4}/0.01$ & 43.64 & 49.77 & 78.77 & 63.91 & \textbf{83.88} & 62.31 & 63.71 \\
$1e^{-4}/0.03$ & 42.57 & 50.2 & 78.61 & \textbf{64.25} & 83.55 & 61.87 & 63.51 \\
$1e^{-4}/0.1$ & 40.11 & 47.95 & 77.98 & 63.4 & 83.37 & 62.19 & 62.5 \\
$1e^{-4}/0.3$ & 35.94 & 44.4 & 78.93 & 62.2 & 82.92 & 62 & 61.07 \\
\hline
$3e^{-5}/0.01$ & \textbf{45.57} & 51.18 & 78.93 & 63.82 & 83.54 & 62.44 & \textbf{64.25} \\
$3e^{-5}/0.03$ & 43.33 & \textbf{51.24} & 78.61 & 64.08 & 83.41 & \textbf{62.61} & 63.88 \\
$3e^{-5}/0.1$ & 41.55 & 49.25 & 78.69 & 62.54 & 83.31 & 62.44 & 62.96 \\
$3e^{-5}/0.3$ & 38.44 & 43.94 & 78.14 & 61.77 & 83.5 & 62.31 & 61.35 \\
\hline
$1e^{-5}/0.01$ & 42.46 & 50.41 & 78.69 & 63.14 & 83.27 & 62.43 & 63.4 \\
$1e^{-5}/0.03$ & 41.58 & 50.35 & \textbf{79.01} & 62.71 & 83.35 & 62.59 & 63.27 \\
$1e^{-5}/0.1$ & 41.17 & 48.63 & 78.69 & 62.54 & 83.38 & 62.35 & 62.79 \\
$1e^{-5}/0.3$ & 39.24 & 44.24 & 78.45 & 62.2 & 83.55 & 62.5 & 61.7 \\
\hline
$3e^{-6}/0.01$ & 39.92 & 48.42 & 78.53 & 62.2 & 83.32 & 62.35 & 62.46 \\
$3e^{-6}/0.03$ & 40.41 & 48.08 & 78.69 & 62.2 & 83.34 & 62.5 & 62.54 \\
$3e^{-6}/0.1$ & 38.63 & 46.3 & 78.22 & 61.95 & 83.39 & 62.58 & 61.85 \\
$3e^{-6}/0.3$ & 38.86 & 43.99 & 77.9 & 61.52 & 83.48 & 62.59 & 61.39 \\
\hline
\end{tabular}
\caption{The impact of different learning rates for SFT on the old HuggingFace Open LLM Leaderboard}
\label{tab:UFT_lr_old_leaderboard}
\end{table*}

\end{document}